# Mining Associated Text and Images with Dual-Wing Harmoniums

**Eric P. Xing**
CALD & LTI
School of Computer Science
Carnegie Mellon University
Pittsburgh, PA 15213

**Rong Yan**
Language Technology Institute
School of Computer Science
Carnegie Mellon University
Pittsburgh, PA 15213

**Alexander G. Hauptmann**
Language Technology Institute
School of Computer Science
Carnegie Mellon University
Pittsburgh, PA 15213

## Abstract

We propose a multi-wing harmonium model for mining multimedia data that extends and improves on earlier models based on two-layer random fields, which capture bidirectional dependencies between hidden topic aspects and observed inputs. This model can be viewed as an undirected counterpart of the two-layer directed models such as LDA for similar tasks, but bears significant difference in inference/learning cost tradeoffs, latent topic representations, and topic mixing mechanisms. In particular, our model facilitates efficient inference and robust topic mixing, and potentially provides high flexibilities in modeling the latent topic spaces. A contrastive divergence and a variational algorithm are derived for learning. We specialized our model to a *dual-wing harmonium* for captioned images, incorporating a multivariate Poisson for word-counts and a multivariate Gaussian for color histogram. We present empirical results on the applications of this model to classification, retrieval and image annotation on news video collections, and we report an extensive comparison with various extant models.

## 1 Introduction

The rapid improvement of processor speed and network systems and the availability of inexpensive massive digital storages have led to a growing demand for modeling and mining data from multiple media sources such as text, images, audio and video (Smeaton and Over, 2003, Hauptmann et al., 2003). To exploit the rich information offered by these heterogeneous features in data mining tasks such as classification, retrieval and image annotation, it is often necessary to model the associated data from multiple sources jointly and/or explore appropriate lower-dimensional latent representations of the originally high-dimensional features.

Numerous approaches for capturing low-dimensional latent representations are available, especially in the context of information retrieval. For example, latent semantic indexing (LSI) (Deerwester et al., 1990) finds a linear transform of word counts into a latent eigenspace of document semantics. The probabilistic latent semantic indexing (pLSI) model (Hofmann, 1999) extends the LSI idea to a probabilistic framework by assuming words to be marginally *iid* samples from a document-specific mixture of word distributions. The mixture of unigrams model (Blei et al., 2003) is a special case of pLSI where each document is associated with only one topic. The latent Dirichlet allocation (LDA) model by Blei et al. (2003) offers a more expressive and generalizable text modeling scheme by associating with each document a unique latent *topic-mixing* vector represented by a random point (under some "prior" distribution) in a simplex. Each word is independently sampled according to different topic draws from the topic mixture, therefore generatively, topic mixing can be achieved at the document level when these words (possibly from different topics) are pooled. (At word level, topic mixing is achieved via marginalization of the latent topic indicators for each word, which in practice depends on the quality of inference computation.) It is straightforward to extend these models to handle multimedia data. For example, van Gemert (2003) applied LSI to capture the joint latent semantic space of text and images. Blei and Jordan (2003) have extended the mixture of unigrams and the LDA model into a Gaussian-multinomial mixture (GM-Mix) and a Gaussian-multinomial LDA (GM-LDA) to model captioned images.

Essentially, all the aforementioned models can be understood as a two-layer directed graphical model (i.e., a Bayesian network) with one layer of hidden units and one layer of input units connected by edges pointing from the hidden layer. The hidden layer can be taken as a representation of the "latent topic aspects" and the input layer corresponds to the observed features of a sample such as a document. Such a Bayesian network formalism offers clear causal semantics and manipulability from a modeling point of view. However, as pointed out by Welling et al. (2004), inference (of the latent topics) in such models can be prohibitively expensive due to the conditional dependencies between all hidden variables. This drawback could seriously affect the model performance in real-time prediction tasks and in EM-based learning. Also worth investigating is the effect of extant representations of topic aspects and the mechanisms of topic mixing. A dominating representation (as used in LDA) for multiple topic aspects is to define it as a point in a *topic simplex*, which can be used as a multinomial parameter vector. As mentioned above, the topic mixing is achieved via repeated draws of topics from this multinomial for each word, or implicitly via marginalization. This scheme enjoys important advantage such as easy sampling and amenability to simple and conjugate topic priors (e.g., Dirichlet). However, as discussed later, it may also be susceptible to poor topic mixing in case of low word-counts and to risks of skewed estimation of topic weights.

In this paper, we present a *multi-wing harmonium* (MWH) model for multimedia data based on a two-layer undirected graphical model called harmonium (Smolensky, 1986, Welling et al., 2004). This model can be viewed as an undirected counterpart of the aforementioned directed aspect models such as LDA, with the following distinctions in topic representation and mixing scheme. Rather than treating the topic vector as a random point in a simplex, we model it as a normally distributed random vector. Rather than mixing topics word by word for text and region by region for image (Blei and Jordan, 2003), we directly model the word counts via Poisson distributions whose rates are determined by the combination of topic aspects and the whole image color histogram via a multivariate Gaussian whose mean is determined similarly. Although it is too early (and indeed unnecessary) to tell which of the two alternative formalisms (i.e., directed or undirected models) are superior for associated text and images, the MWH model does enjoy some unique advantages: 1) inference is fast due to the conditional independence of the hidden units; 2) topic mixing can be achieved by document- and feature-specific combination of aspects rather than via a cumulative effect of single topic draws. We present a contrastive divergence and a variational learning algorithms for model estimation, and we evaluate a specialized *dual-wing harmonium* (DWH) model on classification, retrieval and image annotation on news video collections provided by TRECVID 2003 (Smeaton and Over, 2003) and report an extensive comparison with various extant models.

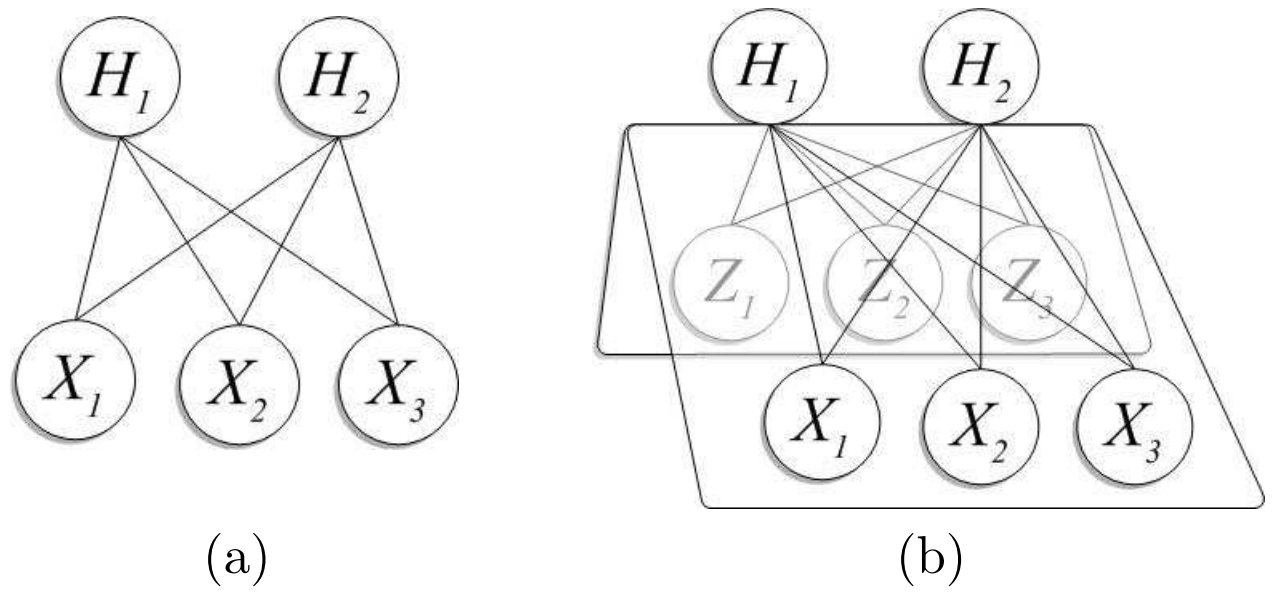

Figure 1: The graphical model representation for (a) a harmonium with 2 hidden units and 3 input units and (b) a multi-wing harmonium with 2 hidden units and 2 wings of input units, where each wing contains 3 input units.

## 2 The model

### 2.1 The basic harmonium model

The harmoniums, which were originally studied by Smolensky (1986) in his *harmony theory*, refer to a family of undirected graphical models defined on complete bipartite graphs containing two layers of nodes (Fig 1a). Let $\mathbf{H} = \{H_j\}$ denote the set of *hidden units* in such a graph, and let $\mathbf{X} = \{X_i\}$ denote the set of *input units*. A harmonium induces a random field

$$p(\mathbf{x}, \mathbf{h}|\theta) = \exp\Big\{\sum_i \theta_i \phi_i(x_i) + \sum_j \theta_j \phi_j(h_j) + \sum_{i<j} \theta_{ij} \phi_{ij}(x_i, h_j) - A(\theta)\Big\},$$

where $\phi_e(\cdot)$ denotes the potential function defined on either a singleton or a connected pair of units (indexed by $e$) in the model, $\theta_e$ denotes the weight of the corresponding potential, and $A(\theta)$ stands for the log-partition function.

The bipartite topology of the harmonium graph suggests that nodes within the same layer are conditionally independent given all nodes of the other opposite layer. This makes possible a very convenient constructive definition of the harmonium distribution based on two between-layer conditional distribution functions $p(\mathbf{x}|\mathbf{h})$ and $p(\mathbf{h}|\mathbf{x})$, both of which factorize over individual units: $p(\mathbf{x}|\mathbf{h}) = \prod_i p(x_i|\mathbf{h})$, $p(\mathbf{h}|\mathbf{x}) = \prod_j p(h_j|\mathbf{x})$. And hence it is semantically simple and easy to design. For simplicity, consider the case where all conditionals are from the exponential family:

$$p(x_i|\mathbf{h}) = \exp\Big\{\sum_a \hat{\theta}_{ia} f_{ia}(x_i) - A_i(\{\hat{\theta}_{ia}\})\Big\}, \quad (1)$$

$$p(h_j|\mathbf{x}) = \exp\Big\{\sum_b \hat{\lambda}_{jb} g_{jb}(h_j) - B_j(\{\hat{\lambda}_{jb}\})\Big\}, \quad (2)$$

where $\{f_{ia}(\cdot)\}$ and $\{g_{jb}(\cdot)\}$ denote the sufficient statistics of variable $x_i$ and $h_j$, respectively; $A_i(\cdot)$ and $B_j(\cdot)$ denote the respective log-partition functions; and the shifted parameters $\hat{\theta}_{ia}$ and $\hat{\lambda}_{jb}$ are defined as,

$$\hat{\theta}_{ia} = \theta_{ia} + \sum_{jb} W_{ia}^{jb} g_{jb}(h_j), \hat{\lambda}_{jb} = \lambda_{jb} + \sum_{ia} W_{ia}^{jb} f_{ia}(x_i),$$

where the shifts are induced by the total couplings between units in the input and hidden layers. Welling et al. (2004) showed that these easily comprehensible and manipulable local conditionals precisely map to the harmonium random fields:

$$p(\mathbf{x}, \mathbf{h}) \propto \exp\Big\{ \sum_{ia} \theta_{ia} f_{ia}(x_i) + \sum_{jb} \lambda_{jb} g_{jb}(h_j) + \sum_{ijab} W_{ia}^{jb} f_{ia}(x_i) g_{jb}(h_j) \Big\}, \quad (3)$$

where $\theta_{ia}, \lambda_{jb}$, and $W_{ia}^{jb}$ are the set of parameters associated with their corresponding potential functions. The log-partition function for this joint probability is not explicitly shown in order to emphasize the difficulty of its estimation. Such a model was referred to as an *exponential family harmonium* (EFH) (Welling et al., 2004). In the sequel, we will take advantage of this bottom-up strategy to construct task-specific harmoniums from easily comprehensible local conditionals.

It can be shown that there is no *marginal* independence for either input or hidden variables in a harmonium. However, an EFH enjoys the advantages of *conditional* independence between hidden variables, which is generally violated in the directed models. This property greatly reduces inference cost. But typically, learning harmonium is more difficult due to the presence of a global partition function.

### 2.2 Multi-wing harmoniums

Although symbolically the hidden and input units in a harmonium are symmetric and thus might appear to preclude causal semantics, the constructive definition described above based on local conditionals between opposite layers indeed provides a vehicle to attribute (bidirectional) causal interpretation of the harmonium structure. Essentially, the hidden units $\mathbf{H}$ can be thought of as the "latent topic aspects" that define how the inputs are generated (conversely, it is also valid to view $\mathbf{H}$ as the predictors resulted from a discriminative model taking the inputs).

In many applications, the input to the model does not have to be from a single source and/or of a homogeneous data type. For example, in a typical multimedia paradigm such as video-stream analysis, the input from a video clip may contain multiple related information such as closed captions, imageries, sound tracks and motion vectors. Assuming that all such inputs are coherent, meaning that they reflect (from different perspectives) the same central theme, then it is natural to model the shared central theme using a set of hidden units, and to group observations from all sources into multiple homogeneous arrays of input units, each corresponding to a single source. This motivates a *multi-wing harmonium* (MWH) (Fig 1b) consisting of multiple canonical harmoniums joint by a shared array of hidden units [1]. Constructing a multi-wing harmonium from a canonical harmonium is straightforward. For example, in case of a dual-wing harmonium, we introduce a second set of input units $\mathbf{z} = \{z_k\}$, which can be related to $\mathbf{h}$ via $p(\mathbf{z}|\mathbf{h}) = \prod_k p(z_k|\mathbf{h})$, where

$$\begin{aligned} p(z_k|\mathbf{h}) &= \exp\big\{ \sum_c \hat{\eta}_{kc} e_{kc}(z_k) - C_k(\{\hat{\eta}_{kc}\}\big\} \quad (4) \\ \hat{\eta}_{kc} &= \eta_{kc} + \sum_{jb} U_{kc}^{jb} g_{jb}(h_j). \end{aligned}$$

Together with Eq. (1), and a slightly modified Eq. (2) that takes into account the influences from both $\mathbf{x}$ and $\mathbf{z}$ by loading the parameters $\hat{\lambda}$ with additional shift

$$\hat{\lambda}_{jb} = \lambda_{jb} + \sum_{ia} W_{ia}^{jb} f_{ia}(x_i) + \sum_{kc} U_{kc}^{jb} e_{kc}(z_k),$$

where $\{U_{kc}^{jb}\}$ captures the couplings between the hidden units and the $\mathbf{z}$ inputs, we obtain the dual-wing harmonium random fields:

$$\begin{aligned} & p(\mathbf{x}, \mathbf{z}, \mathbf{h}) \propto \\ & \exp\Big\{ \sum_{ia} \theta_{ia} f_{ia}(x_i) + \sum_{kc} \eta_{kc} e_{kc}(z_k) + \sum_{jb} \lambda_{jb} g_{jb}(h_j) \\ & + \sum_{ijab} W_{ia}^{jb} f_{ia}(x_i) g_{jb}(h_j) + \sum_{kjcb} U_{kc}^{jb} e_{kc}(z_k) g_{jb}(h_j) \Big\}. (5) \end{aligned}$$

This construction maintains the conditional independence between hidden variables given inputs and hence ensures the efficiency of inference (once the model is parameterized). Note that $\mathbf{x}, \mathbf{z}$ are marginally dependent to each other, as can be quickly verified from the bipartite graph structure. This enables the application of inferring the features of one source from anther source, such as automatic image annotation (Blei and Jordan, 2003) which attempts to infer related words from a given image.

[1] Note that modulo descriptive semantics, the multi-wing harmonium is mathematically the same as the canonical harmonium. But categorizing the input units into multiple homogeneous arrays (i.e., wings) makes it explicit that the underlying local conditionals of each wing can be designed separately to reflect unique characteristics of different sources, as we demonstrate in the sequel.

### 2.3 A dual-wing harmonium for text and images

To model video streams, which contain both text and image information, in the following we outline a *dual-wing harmonium* (DWH) model based on a text submodel and an image submodel using the modular constructive technique described above. Following the tradition of the bag-of-word model for texts, we model each document with its word-count profile. Instead of using a continuous surrogate of the discrete counts (as done in a mixture of Gaussian setting), or assuming that the counts are accumulated from independent draws from multinomial distributions (as done in LDA), we assume that the latent topic aspects associated with each document directly determine the expected rate of each word in a document. This is done by specifying a Poisson distribution for the observed count of each word in the document. As discussed later, this text model has the key difference from a multinomial model in that topic mixing is achieved directly in the distribution of word rates determined by the document-specific *combination of topic aspects*, rather than via an additive effect of multiple *single topic draws* of the same word or via marginalization of the latent topic of each word. In our conditional Poisson model for word counts, topic mixing is still stable and robust even when a word appears only once or a few time in a document (which is typical in video captions). Whereas in the multinomial word model, single word instance can only come from a single topic and is thus unable to enjoy topic mixing directly. Specifically, our text model is as follows. For each word $i \in \{1, \ldots, M\}$, its rate $x_i$ is distributed as:

$$p(x_i|\mathbf{h}) = \text{Poisson}(x_i|\exp(\alpha_i + \sum_j h_j W_{ij})), \quad (6)$$

where the shift of the Poisson rate $\alpha_i$ is induced by a weighted combination of the latent topic aspects $\mathbf{h}$.

For the image inputs, we adopt a color-histogram representation of the images, which is typically modeled by a Gaussian distribution:

$$p(z_k|\mathbf{h}) = \mathcal{N}(z_k|\sigma_k^2(\beta_k + \sum_j h_j U_{kj}), \sigma_k^2), \quad (7)$$

where $z_k$ denote the $k^{th}$ bin value of the color histogram for an image, and the shift of the mean is again determined by a (different) weighted combination of the (same) latent topic aspects $\mathbf{h}$. To guarantee the conditional probabilities of image inputs and hidden units are consistent (i.e., resulting from the same joint probability), we scale the mean of $p(z_k|h)$ by $\sigma_k^2$.

Finally for the hidden units $\mathbf{h}$ which represent the latent topic aspects of the input data, following Welling et al. (2004), we assume that each aspect admits a conditional unit-variance Gaussian distribution whose mean is determined by a weighted combination of the observed word counts and the color histogram.

$$p(h_j|\mathbf{x}, \mathbf{z}) = \mathcal{N}(h_j|\sum_i x_i W_{ij} + \sum_k z_k U_{kj}, 1). (8)$$

Note that unlike several multinomial-based aspect models, which represent the conjoint vector of topics that contribute to the textual or image content of the study object as a point in a simplex, here we represent the topic vector as a random point in Euclidean space (we will discuss the benefit of such an approach later along with our experimental results).

Putting everything together, it can be shown that the marginal distribution of the input units is:

$$p(\mathbf{x}, \mathbf{z}) \propto \exp\Big\{ \sum_i \alpha_i x_i - \sum_i \log \Gamma(x_i) + \sum_k \beta_k z_k - \frac{1}{2}\sum_k \frac{z_k^2}{\sigma_k^2} + \frac{1}{2}\sum_j (\sum_i W_{ij} x_i + \sum_k U_{kj} z_k)^2 \Big\}. (9)$$

Note that in Eq. (8) we define the variance of the hidden variables given the inputs to be one to simplify the parameter estimation. Introducing a covariance matrix $\Sigma$ in the $p(h_j|x_i, z_k)$ can offer additional freedom for the joint probability, but it would not lead to more general representations in terms of the marginal probability (Welling et al., 2004).

## 3 Learning and Inference

Given an iid sample $\mathcal{X} = \{\mathbf{x}_n, \mathbf{z}_n\}_{n=1}^N$, we can estimate the parameters of a harmonium under a maximum likelihood objective by gradient ascent. The learning rules (i.e., the gradients) can be obtained by taking derivative of the log-likelihood of the sample defined by Eq. (9) with respect to the model parameters:

$$\delta\alpha_i = \langle x_i \rangle_{\tilde{p}} - \langle x_i \rangle_p \ , \quad \delta\beta_k = \langle z_k \rangle_{\tilde{p}} - \langle z_k \rangle_p \ ,$$
$$\delta(\sigma_k^{-1}) = \langle z_k^2 \sigma_k^{-1} \rangle_{\tilde{p}} - \langle z_k^2 \sigma_k^{-1} \rangle_p,$$
$$\delta W_{ij} = \langle x_i h'_j \rangle_{\tilde{p}} - \langle x_i h'_j \rangle_p, \delta U_{ij} = \langle z_k h'_j \rangle_{\tilde{p}} - \langle z_k h'_j \rangle_p,$$

where $\langle \cdot \rangle_p$ denotes an expectation w.r.t. distribution $p$, $\tilde{p}(x) = \frac{1}{N}\sum_n \delta(x - x_n)$ denotes the empirical data distribution, $p(\cdot)$ stands for the model distribution (i.e., the harmonium random fields), and $h'_j$ stands for $\sum_i W_{ij} x_i + \sum_k U_{kj} z_k$. Note that due to the presence of the partition function $Z$ in $p$, computing the second expectation is usually intractable. In the following, we briefly describe two algorithms for approximate gradient ascent learning.

### 3.1 Contrastive divergence

Instead of doing an exact gradient ascent using the learning rules stated above, following Welling et al. (2004), we can use the contrastive divergence (CD) (Hinton, 2002, Welling and Hinton, 2001) to approximate the learning rules. CD approximates the intractable model distribution using a single or a few iterations of Gibbs sampling, and is therefore highly efficient. But since the diagnosis of convergence is not straightforward, we run the CD learner up to a fixed number of iterations in our experiments.

### 3.2 Variational approximation

Alternatively, we can use a variational approximation to the model distribution $p$. Specifically, we use a *generalized mean field* (GMF) approximation to the intractable harmonium random field, which takes a factorized form as the product of all singleton marginals over the variables (Xing et al., 2003) (indeed this is the simplest GMF which reduces to the standard mean field scheme (Peterson and Anderson, 1987), but upgrades to better GMF approximations are straightforward):

$$q(\mathbf{x}, \mathbf{z}, \mathbf{h}) = \prod_i q(x_i|\nu_i) \prod_k q(z_k|\mu_k, \sigma_k) \prod_j q(h_j|\gamma_j), \quad (10)$$

where $q(x_i|\nu_i)$ is a Poisson distribution with mean $\nu_i$, $q(z_k|\mu_k, \sigma_k)$ is a Gaussian distribution with mean $\mu_k$ and variance $\sigma_k$, and $q(h_j|\gamma_j)$ is a Gaussian distribution with mean $\gamma_j$ and variance 1. According the GMF theorem (Theorem 1 in (Xing et al., 2003)), we have the following fixed point equations for the GMF approximation:

$$\gamma_j = \sum_i W_{ij}\nu_i + \sum_k U_{kj}\mu_k, \quad (11)$$

$$\mu_k = \sigma_k^2(\beta_k + \sum_j U_{kj}\gamma_j), \quad (12)$$

$$\nu_i = \exp(\alpha_i + \sum_j W_{ij}\gamma_j), \quad (13)$$

which minimize the KL divergence between $q$ and $p$. Upon convergence, we use the resulting $q$ as a surrogate to the original harmonium random fields $p$, and compute the expected gradients for parameter updating. Typically, this scheme is more efficient than CD, but usually leads to less accurate results. Note that although the dependency between variables are fully decoupled in the Eq. (10), GMF can offer tighter approximations based on cluster marginals associated with a block partition on the variables. These approximations will be explored in the future work.

### 3.3 Inferring words from images

A useful application for a multi-wing harmonium is to find the marginal (i.e., after integrating out the latent topic aspects) dependency relationships between various types of inputs, such as inferring the most related words $\{x_i\}$ from a given image $\mathbf{z}$. Unfortunately, no analytical solutions are available for computing the conditional probability $p(x_i|\mathbf{z})$, but the variational approximation to the harmonium random fields described above provides a tractable approach to approximate $p(x_i|\mathbf{z})$. It is easy to see from Eq. (10) that the variational approximation to the conditional of interest, e.g., $p(x_i|\mathbf{z})$, is simply the GMF singleton marginal $q(x_i|\nu_i)$, which readily comes out of the fixed point iterations (i.e., Eqs. (11-13) performed during training. We can select the top-ranked words corresponding to an image $\mathbf{z}$ by sorting all words in the descending order of $q(x_i|\nu_i)$.

## 4 Experiments

We compiled our dataset by sampling the TRECVID'03 news video collection (Smeaton and Over, 2003). The video clips are segmented into multiple video shots which can be viewed as a "document" or a "training/testing example". Based on the common annotations (Smeaton and Over, 2003), 1078 video shots belonging to five categories were selected in our experiments, i.e. *Airplane Scene, Basketball Scene, Weather News, Baseball Scene, Hockey Scene*, each shot is associated with one category. For each shot, we extract 1894 binary word presence features from the associated closed captions and 166 dimensional color correlogram in $HSV$ color space (Hauptmann et al., 2003) from the key image. In order to balance the contributions between both features types, image features are linearly normalized to guarantee the sums of text and image features are equal in each shot.

By default, DWH is trained via contrastive divergence with up to 1000 steps of gradient ascent. For comparison, we also used a variational scheme for training. To mitigate the issue of "identifiability" (Welling et al., 2004) that allows multiple parameter configurations to share the same marginal likelihood, the initial estimations of parameters $W$ and $U$ in DWH were determined by a SVD on the design matrix of text/images features over shots. We do not strongly emphasize this issue because in our analysis DWH is not mainly used to directly capture the exact semantics of the latent factors underlying the data space (although empirically we found that our simplistic strategy still yields reasonable semantics as seen in §4.1). In order to achieve semantically more accurate and informative latent factor representations, we can apply a subsequent clustering procedure on the lower-dimensional representations provided by DWH.

For the purpose of comparison, we also implemented three related models—LSI, GM-Mix and GM-LDA. The parameters of GM-Mix and GM-LDA were ob-tained using EM. We infer the latent topic captured by GM-Mix using the conditional probabilities of hidden variables $p(\mathbf{h}|\mathbf{x}, \mathbf{z})$ and those by GM-LDA based on variational Dirichlet posteriors of the topic weights. For simplicity, we omit details, but see (Blei and Jordan, 2003) for more details.

### 4.1 Illustrative examples of latent topics

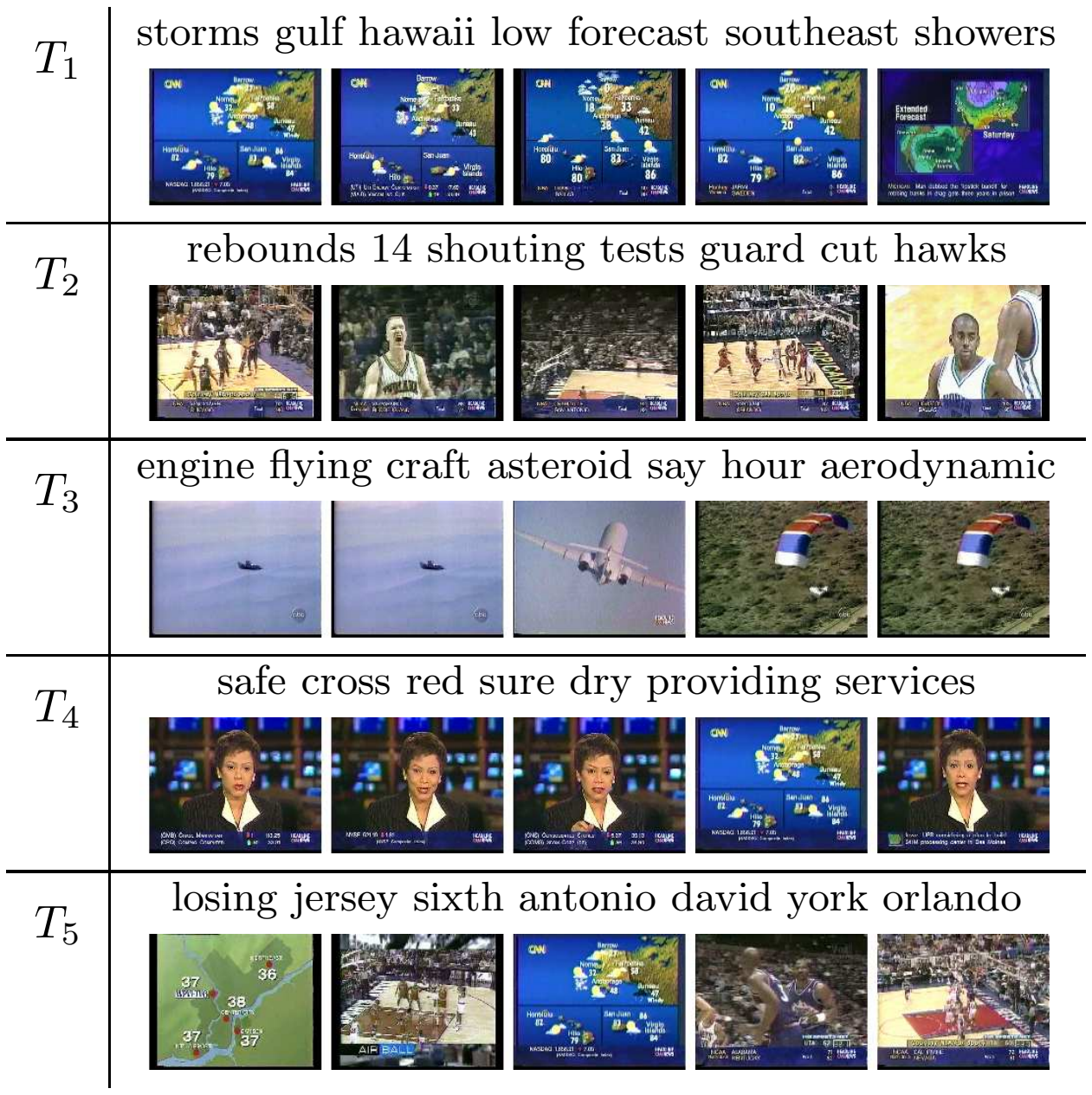

Figure 2: An illustration of 5 latent topics. Each topic is shown with the top 10 words and the top 5 images extracted from the most related video shots.

As an empirical demonstration of DWH's ability to automatically discover meaningful latent topics from both text and images, we illustrate 5 latent topics out of the 20 topics learned by DWH in Fig 2. Each topic is described by the top 10 words and the top 5 key images associated with the video shots that provide the highest conditional probabilities on the latent topic. The examples of the first three topics clearly correspond to the scenes of Weather News, Basketball and Airplane, respectively, which is clustered based on the evidence from both words and images. The fourth topic is kind of narrow, which captures the scene of the same anchorperson from CNN Headline News. This topic is likely to be generated from the evidence primarily from image similarities. The last topic illustrates some interesting patterns discovered by DWH. At a first sight, these shots appear to follow a degenerated theme because they cover different scenes including both basketball and weather news. But after reviewing the associated closed-captions, we found that these shots do share some common aspects. First, they all mention similar place names and numbers in the transcript such as "(New) York", "(New) Jersey" and "sixth". Second, both the weather news and basketball reports uses the same terms in reporting, such as "losing". Apparently, DWH discovers the last topic mainly based on the word similarities.

### 4.2 Classification, retrieval and image annotation

Now we report a series of experiments demonstrating the predictive power of the lower-dimensional representations produced by DWH. We evaluate DWH on three important tasks in image and text mining. Typically, we set the dimensions of the latent topic aspects to be less than 50 (compared to the original ∼2000 dimension for text and 166 for image). It is interesting to examine what can be achieved even with such a significant rate of information loss.

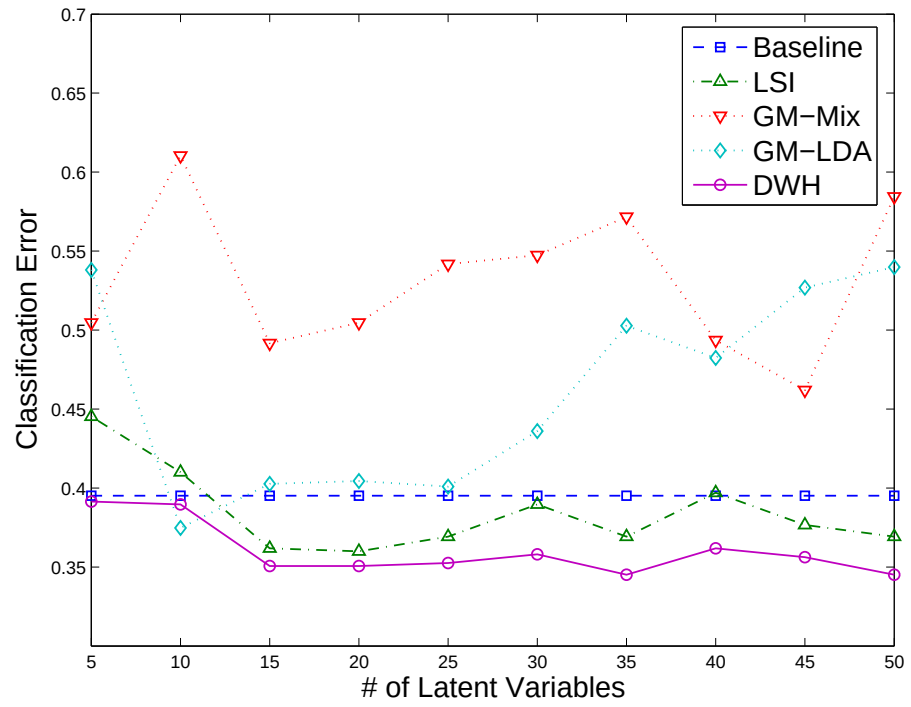

(a)

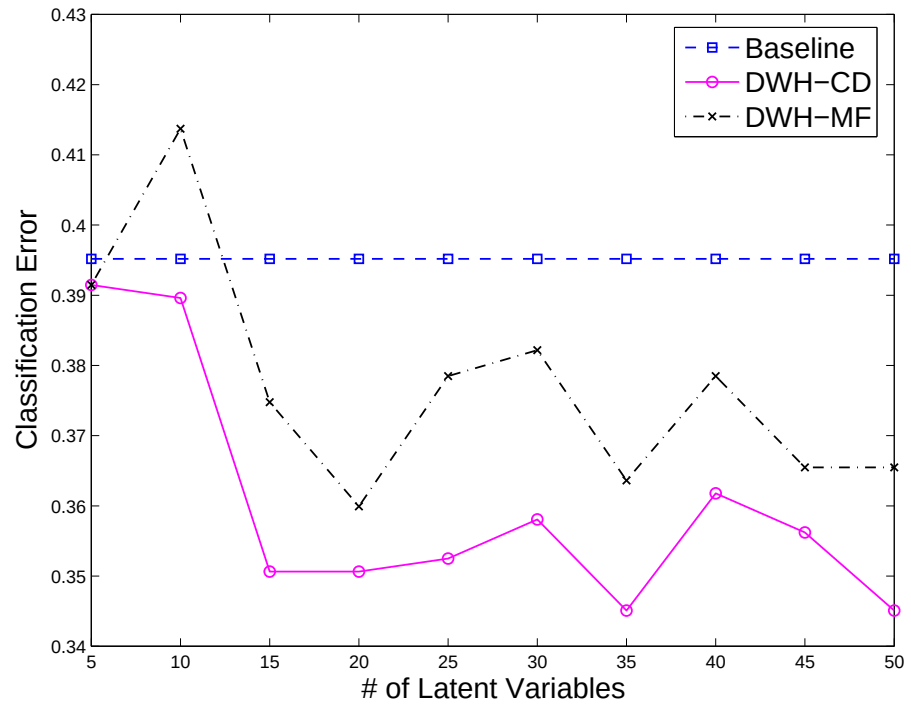

(b)

Figure 3: Classification errors versus the dimensions of latent aspects (i.e., number of hidden units). (a) A comparison of five approaches including baseline, LSI, GM-Mix, GM-LDA and DWH; (b) A comparison of DWHs trained via contrastive divergence and variational learning.

First we evaluate the performance of DWH (and other algorithms) on classifying testing examples into one of the pre-defined categories. For each algorithm, the parameters are estimated using all data, ignoring their true class labels. Once the models are learned, we use them to project every example into a lower-dimensional latent topic space. Then we split the data evenly into a training set and a testing set, use the SVM$^{\text{Light}}$ package to learn a support vector machine (SVM) on the training data, and predict on the testing data.

Fig 3(a) compares the classification errors of five different models—baseline, LSI, GM-Mix, GM-LDA and DWH—obtained at different dimensions (ranging from 5 to 50) of the latent topic aspects (i.e., the number of hidden units in case of DWH). The baseline approach keeps all available features for both training and testing. We find that even with a large dimensionality reduction, DWH almost always achieve a significantly lower classification error than the baseline. It also outperforms LSI with a good margin under the same topic dimension. We believe that this may be partially explained by the the arguably better assumptions adopted by DWH on modeling text/image features. Surprisingly, GM-Mix produces a considerably worse performance than the baseline because the modeling power of GM-Mix is too limited to capture multiple latent topics for each text/image pair. Too much information is eliminated in GM-Mix's representations because the posterior distribution is usually peaked at one latent topic. Compared to GM-Mix, GM-LDA offers more flexibilities in modeling associated text/images and indeed it is (slightly) superior to all other models when latent aspect dimension is set to be 10. But it appears that GM-LDA may have suffered from overfitting or a low-dimensionality bias as its error curve rises significantly in higher-dimensional latent space. In contrast, we observe that the performances of LSI and DWH are relatively stable over a wide range of dimensions, which may reflect the robustness and expressiveness of their representation schemes for the latent aspects (i.e., as Gaussian variables rather Dirichlet variables).

In Fig 3(b) we compare the performance of DWH learned using CD and variational methods. CD achieves lower error than variational methods, likely because the latter approach uses a fully factorized distribution to approximate the true distribution whereas the former uses a Monte Carlo approximation. Running the Gibbs samplers to full convergence might provide even higher accuracy, but this is not practical because it would take an unaffordable amount of time to conver even a small dataset.

Retrieval is another standard task for mining associated text and images, of which the goal is to rank video shots in a descending order of the relevance to a given query. In this experiment, we use the same setting for training and testing as described above. Specifically, we consider each example in the testing set as a query and compute the standard cosine similarity to rank the training examples. An example is relevant to the query if it belongs to the same category. The retrieval results were averaged over all the testing examples. We use the precision, recall and (non-interpolated) average precision (AP) (Smeaton and Over, 2003) over a set of retrieved documents as the measures of retrieval effectiveness. Roughly speaking, average precision corresponds to the area under the precision-recall curve. Fig 4 compares the average precisions of DWH and three other models, as well as a baseline using TF-IDF weighting without any dimension reduction. Again, performances are reported over dimensions of latent aspects ranging from 5 to 50. Similar to the classification results, DWH consistently outperforms LSI and the baseline method. GM-Mix is often worse than other approaches. The performance of GM-LDA peaks at the point of 10 latent variables but goes down afterwords. The precision-recall curve for a fixed aspect dimension (i.e., 20) is shown in Fig 5.

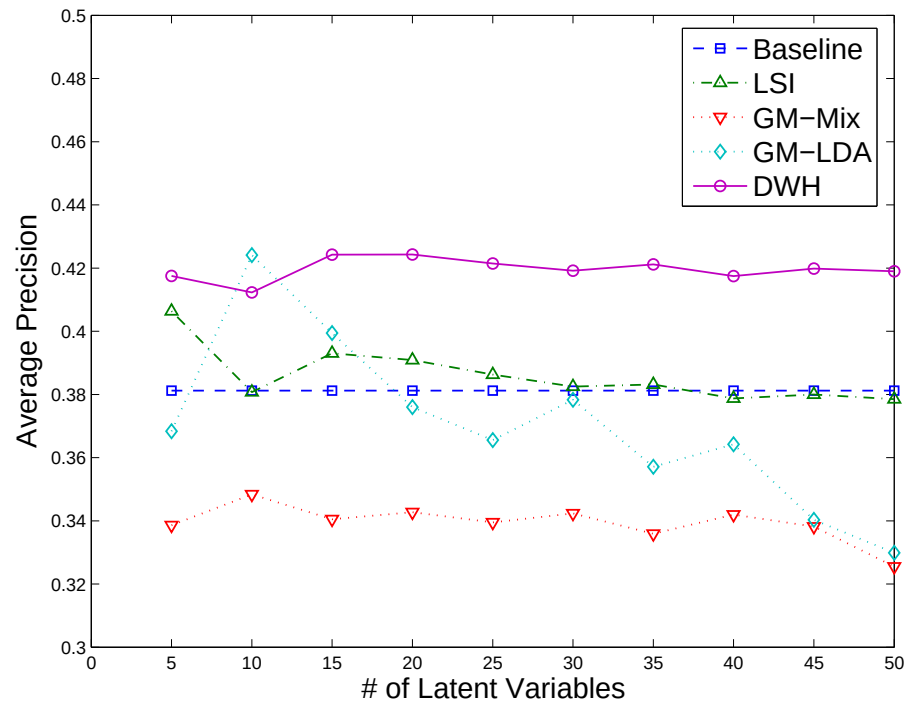

Figure 4: The average precision (AP) of text/image retrieval versus the dimensions of latent aspects. Five approaches are compared.

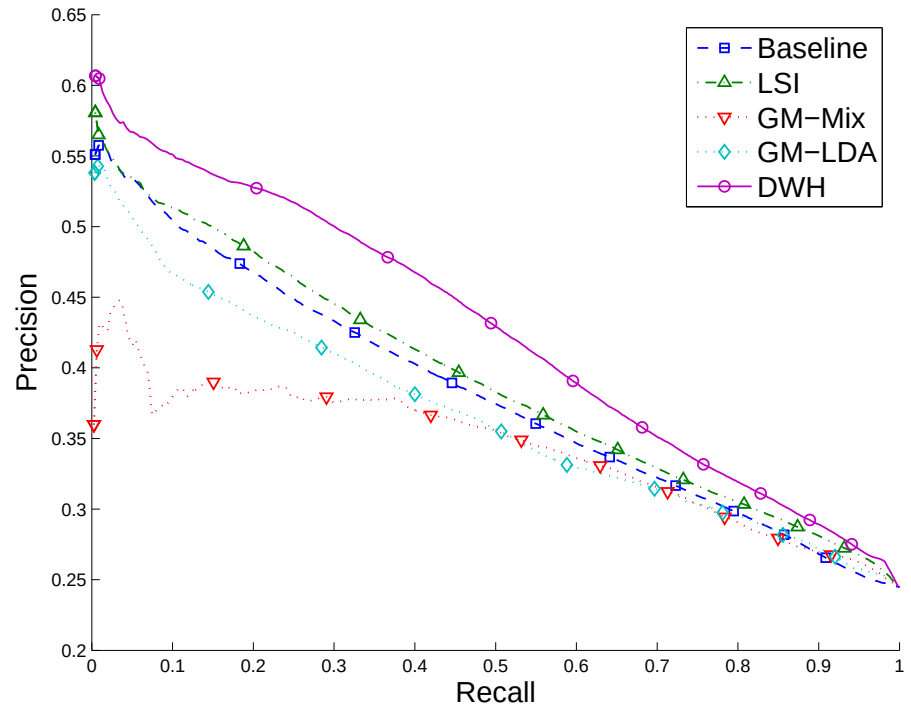

Figure 5: The precision-recall curves for five approaches under 20 latent aspects.

Finally we examine the performances of DWH on image annotation, i.e., annotating an unseen image with the most relevant words. The training and testing set are constructed as before. For each given image, we rank all possible words in the vocabulary based on the descending order of the conditional probabilities of words given the image $P(\mathbf{x}|\mathbf{z})$. The closed captions associated with the given images are treated as the ground truth. Since LSI cannot handle the image annotation explicitly, only GM-Mix, GM-LDA and DWH are considered in this task. Fig 6 shows a comparison of their performances under different aspect dimensions. As noted by Blei and Jordan (2003), GM-LDA performs poorer than GM-Mix for image annotation because it is too flexible and allows caption words to be generated by the factors that is not related to the image region (Blei and Jordan, 2003) [2]. DWH can achieve better average precisions than GM-Mix especially when more latent dimensions remain, although the gap is not as significant as in other tasks.

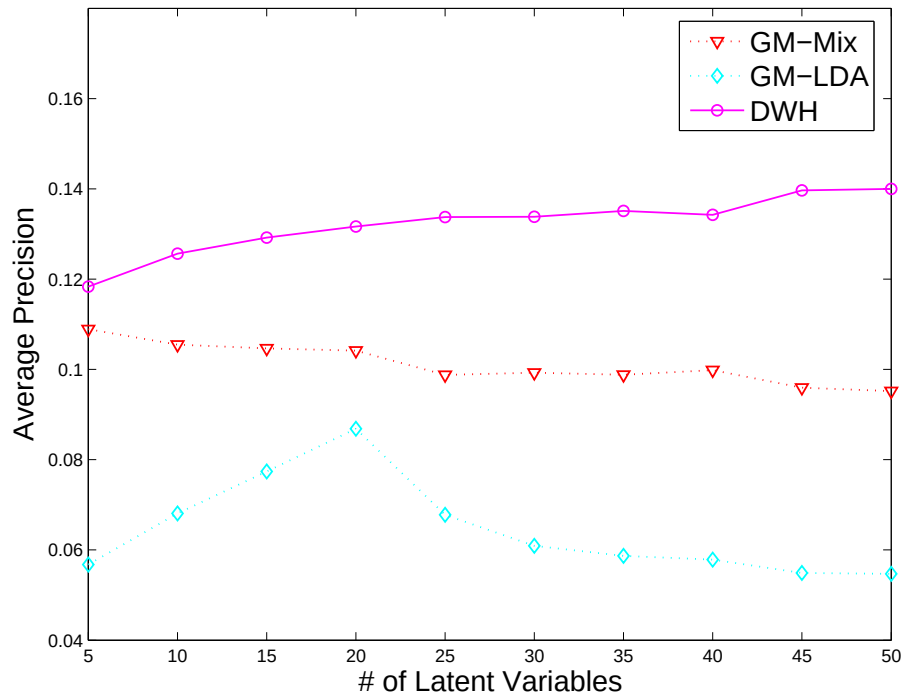

Figure 6: The average precision of the image annotation under different dimensions of latent aspects. Three approaches are compared (GM-Mix, GM-LDA and DWH).

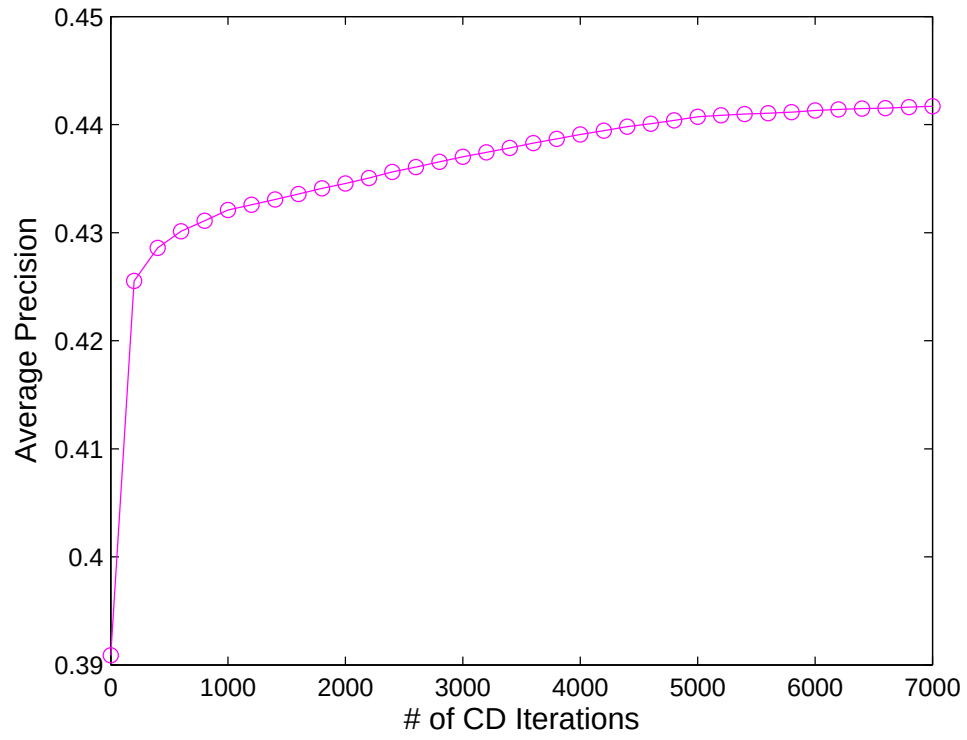

Figure 7: The average precision(AP) of retrieval when learning DWH using contrastive divergence versus the number of iterations which grows from 0 to 7000 (plotted at an interval of 200).

### 4.3 More on overfitting and CD learning

A legitimate question over the results reported here is that whether the apparent superiority of DWH is due to the advantage of the model itself or due to a nice property of the CD learning algorithm— it somehow prevents overfitting because it is based on a brief sampling procedure in combination with learning in mini-batches (personal communication with Welling). (This protection tends to diminish when the training is tuned and prolonged to push closer to the optimum of the CD-objective.) In order to verify whether DWH is subject to overfitting when it comes close to the optimum of CD objective function, Fig 7 shows the average retrieval precision of DWH versus the number of iterations for contrastive divergence. As can be seen, the learning curve keeps moving up after the point of 1000 iterations (the default condition for all our experiments). Although detecting the convergence of contrastive divergence is still an unsolved problem, the trend in Fig 7 suggests that DWH is fairly robust (i.e., does not overfit) even when it is trained longer and getting closer to the optimal likelihood. Similar conclusion is also suggested by Fig 3(b), in which the performance of DWH due to variational inference (which is also used for LDA) still dominates over other models. Thus we believe that the superior performance of DWH in our analysis is mainly due to the model, not the inference/learning algorithm.

[2] A correlation-LDA model was proposed to reduce full exchangeability between word features and image features, thereby couples word distributions to certain regions in the image (Blei and Jordan, 2003). But since as of now, DWH has not exploited region-specific features of images, we feel that a comparison to GM-Mix and GM-LDA is more fare. We intend to construct a DWH based on regional color histogram and compare it with correlation-LDA in the full paper.

## 5 Conclusions and Discussions

We have presented a variant of an undirected graphical model known as the harmonium for modeling multimedia data. In this model, the topics are represented by a multivariate Gaussian variable, and inputs of various types are modeled by feature-specific conditional distributions—i.e., a multivariate Poisson distribution for word counts and a multivariate Gaussian for color histogram—determined by the conjoint effects of all topic aspects, thereby achieving topic mixing. Despite its undirected nature, our model, termed a dual-wing harmonium, can be defined constructively from a bidirectional generation process with intuitive causal semantics.

The probabilistic structure of DWH enables efficient inference of the latent topics, at the cost of a substantially more demanding learning task. But since learning can be usually done offline, we consider this tradeoff acceptable, especially for tasks involving real-time predictions such as on-line image annotation. The topic representation/distribution and the topic mixing mechanism in DWH is rather different from those of several extant models such as LDA, and as a result offer very different utilities compared to those models. In particular, the multivariate Poisson-based text model arguably allows a more robust mixing of multiple topic effects on each word. For example, in the extreme cases of having only one occurrence for some key word, DWH still facilitates stable topic mixing because the Poisson word rate is determined directly by a linear combination of all topic aspects (in this regard DWH is similar to the EFH model by Welling et al. (2004) which employs a softmax for each count value and requires a much higher parameterization cost). Whereas in a multinomial text model, a generative process would attribute such words with a single topic, and mixing has to be achieved via marginalization of the latent word topic indicators—an expensive inference step often approximated via Monte Carlo sampling or variational methods. We also suspect that for very large document, the likelihood surface of a multinomial text model would have many local optima due to the prevalence of non-signature words competing for probability mass, whereas in DWH the effect of the total word count can be naturally buffered across all topic aspects via a shared scaling factor, which may make the rates of the Poisson test model less sensitive to spurious wold counts. At this point, it is still unclear whether such effects actually play a significant role in various prediction or retrieval tasks and we intend to investigate further with more thorough experiments (note that it is pointless to merely compare the perplexities under the two models because they use rather different principles to define the probability mass function). It also appears that the Poisson word model can be more easily combined with the models for other sources (e.g., multivariate Gaussian for color histogram) without dominating or being dominated in terms of the contribution of probability mass, whereas a multinomial-Gaussian combination may go off-balance (McCallum and Nigam, 1998).

It is also interesting to explore the difference between topic representations in an unconstrained continuous space and in a simplex. The later enjoys very natural word-frequency interpretation and a number of computational and modeling advantages (Blei et al., 2003). But specifying or training an appropriate Dirichlet prior for this representation can be tricky. Representing topic aspects as a Gaussian variable is more flexible, even allowing certain aspects to be negative. Whether such flexibilities are semantically sensible is not clear and debatable, but they could be indeed useful and convenient in various model extensions such as temporal models for topic evolution and more elaborated Bayesian models of topic distributions. It is possible to manipulate the simplex representation to achieve similar effects (i.e., using a log-normal model over the simplex), and we believe further explorations of different formalisms will lead to new innovations and insights.